\documentclass[conference]{IEEEtran}
\IEEEoverridecommandlockouts
\usepackage{cite}
\usepackage{amsmath,amssymb,amsfonts}
\usepackage{algorithmic}
\usepackage{graphicx}
\usepackage{textcomp}
\usepackage{xcolor}
\usepackage{hyperref}
\def\BibTeX{{\rm B\kern-.05em{\sc i\kern-.025em b}\kern-.08em
    T\kern-.1667em\lower.7ex\hbox{E}\kern-.125emX}}
\begin{document}

\title{Blood Pressure Estimation from PPG: A Comparative Study of Direct and ECG-Mediated Deep Learning Pipelines
}

\author{
\IEEEauthorblockN{Bo Wu}
\IEEEauthorblockA{
\textit{Department of Informatics and Networked Systems} \\
\textit{University of Pittsburgh} \\
Pittsburgh, USA \\
bow36@pitt.edu
}
\and
\IEEEauthorblockN{Haoling Wang}
\IEEEauthorblockA{
\textit{Department of Biostatistics and Health Data Science} \\
\textit{University of Pittsburgh} \\
Pittsburgh, USA \\
haw222@pitt.edu
}
\and
\IEEEauthorblockN{Zhuodiao Kuang}
\IEEEauthorblockA{
\textit{Department of Biostatistics and Health Data Science} \\
\textit{University of Pittsburgh} \\
Pittsburgh, USA \\
zhk22@pitt.edu
}
\and
\IEEEauthorblockN{Kateryna Shapovalenko}
\IEEEauthorblockA{
\textit{Language Technologies Institute} \\
\textit{Carnegie Mellon University} \\
Pittsburgh, USA \\
kshapova@alumni.cmu.edu
}
}

\maketitle

\begin{abstract}
Continuous cuffless blood pressure (BP) monitoring is essential for connected health systems and wearable devices, enabling early detection, longitudinal tracking, and personalized management of cardiovascular disease. Many prior approaches attempt to estimate BP indirectly by reconstructing electrocardiography (ECG) from photoplethysmography (PPG), assuming ECG provides a stronger physiological link to BP. However, ECG sensing is less accessible in wearable settings and may introduce unnecessary complexity.

In this work, we first perform a large-scale physiological correlation analysis on the MIMIC-III waveform database, revealing that PPG exhibits substantially stronger coupling with arterial blood pressure (ABP) ($|r|=0.247$, $p<0.001$) than ECG does ($r=0.018$, $p=0.187$), challenging the assumption that ECG provides a superior intermediate representation. Motivated by this insight, we conduct a systematic comparison between direct PPG-to-BP prediction and ECG-mediated pipelines using multiple state-of-the-art deep learning models.

Across 1.74M segments from 3,127 patients, direct PPG-to-BP prediction achieves British Hypertension Society Grade A performance ($\mathrm{MAE}_{\mathrm{SBP}} = 4.82 mmHg$, $\mathrm{MAE}_{\mathrm{DBP}} = 4.31 mmHg$), outperforming all ECG-mediated approaches, which achieve only Grade B accuracy.

Our findings suggest that accurate continuous BP monitoring can be achieved directly from wearable PPG signals, enabling simpler, more efficient pipelines for real-world connected health systems.
\end{abstract}


\section{Introduction}

Cardiovascular diseases cause 17.9 million deaths annually, with hypertension affecting 1.3 billion people worldwide. Continuous blood pressure monitoring is critical for early detection and management, yet traditional cuff-based measurements are intermittent and uncomfortable. Photoplethysmography (PPG) sensors offer a wearable-friendly and widely accessible solution, but achieving clinical-grade accuracy remains challenging. Many recent approaches attempt to improve performance by reconstructing electrocardiography (ECG) as an intermediate representation, assuming ECG provides a stronger physiological link to blood pressure. This raises a fundamental question: does generating synthetic ECG actually improve BP estimation compared to predicting BP directly from PPG?

Our research objectives are: (1) systematically evaluate whether ECG-mediated pipelines improve BP estimation versus direct PPG-to-BP prediction; (2) benchmark multiple state-of-the-art generative architectures for PPG-to-ECG translation within a unified comparative framework; and (3) determine whether direct PPG-based approaches can achieve clinical-grade accuracy meeting BHS Grade A standards (MAE $\leq$ 5 mmHg, SD $\leq$ 8 mmHg). Answering these questions provides critical guidance for the design of wearable connected health systems enabling continuous, comfortable, and scalable cardiovascular monitoring.

\section{Related Work}

Early cuffless blood pressure estimation approaches relied on hand-crafted features derived from physiological signals such as Pulse Transit Time (PTT), computed from PPG and ECG. These methods require calibration and often exhibit limited robustness and generalization, particularly in wearable and real-world settings. 

Deep learning methods improved performance by learning features directly from raw signals without manual engineering. For example, Kachuee et al. (2018) \cite{kachuee2018} used LSTM models to estimate BP from PPG and ECG, achieving 7--9 mmHg MAE, while El-Basyouni et al. (2019) \cite{elbasyouni2019} combined ResNet and GRU architectures to achieve approximately 6 mmHg MAE.

More recently, generative models have been proposed to reconstruct ECG from PPG as an intermediate representation for BP estimation. Ji and Zhou (2024) \cite{ji2024generative} demonstrated that generative approaches can successfully synthesize physiologically realistic ECG signals from wearable PPG. However, these pipelines introduce additional model complexity and computational overhead.

To our knowledge, no prior work has conducted a systematic, apples-to-apples comparison between direct PPG-based BP prediction and ECG-mediated pipelines within a unified experimental framework.

\section{Data}

\subsection{MIMIC-III Database}

We use MIMIC-III Waveform Database \cite{johnson2016mimic} containing synchronized ECG, PPG, and ABP at 125 Hz from 3,127 ICU patients. ABP provides a gold-standard ground truth for BP values.

\subsection{Preprocessing Pipeline}

Our five-stage pipeline includes: (1) Signal extraction and temporal alignment; (2) Quality control (SNR $>$ 15 dB, valid BP ranges: SBP 80-200 mmHg, DBP 40-120 mmHg); (3) Segmentation into 256-sample windows (~2 sec) with 50\% overlap; (4) Per-segment z-score normalization; (5) BP extraction (SBP: maximum, DBP: minimum). 

Final dataset: 1,743,892 training, 218,234 validation, 217,456 test segments (94.2\% pass quality control). Optionally, we also compare temporal (original data) vs spectral features.

To obtain spectral features, we transform synchronized ECG, PPG, and ABP signals from the time domain to the frequency domain using a real-valued Fast Fourier Transform (rFFT), which produces a one-sided complex spectrum preserving signal magnitude and phase. Because cardiovascular dynamics are primarily concentrated in low-frequency bands, we retain frequency components between 0 and 20 Hz and discard higher-frequency noise.

For each retained frequency bin, we store both real and imaginary components, enabling full signal reconstruction via inverse rFFT if needed. These spectral representations provide a frequency-domain view of physiological dynamics and are used as inputs for downstream model training and evaluation.

\subsection{Correlation Analysis}

We first analyze pairwise correlations between PPG, ECG, and ABP to assess their physiological relationships (Table \ref{tab:correlation}). PPG shows substantially stronger correlation with ABP ($|r| = 0.247$, $p < 0.001$) compared to ECG ($r = 0.018$, $p = 0.187$), indicating that PPG captures more BP-relevant information. This observation motivates our systematic comparison between direct PPG-based prediction and ECG-mediated pipelines.

\begin{table}[htbp]
\caption{Pairwise Correlation of Physiological Signals}
\begin{center}
\begin{tabular}{|c|c|c|c|}
\hline
\textbf{Signal Pair} & \textbf{Pearson $r$} & \textbf{$p$-value} & \textbf{Interpretation} \\
\hline
PPG $\leftrightarrow$ ABP & -0.247 & $<0.001$ & Moderate negative \\
\hline
PPG $\leftrightarrow$ ECG & 0.021 & 0.132 & Very weak \\
\hline
ECG $\leftrightarrow$ ABP & 0.018 & 0.187 & Very weak \\
\hline
\end{tabular}
\label{tab:correlation}
\end{center}
\end{table}

\section{Model Description}

\subsection{Experimental Framework}

We evaluate two pipelines for estimating systolic and diastolic blood pressure (SBP/DBP) from wearable PPG signals using arterial blood pressure (ABP) as ground-truth supervision. The \textbf{Direct Pipeline (PPG$\rightarrow$BP)} predicts BP directly from PPG using a BiLSTM model. The \textbf{ECG-Mediated Pipelines (PPG$\rightarrow$ECG$\rightarrow$BP)} first reconstruct ECG from PPG using a generator, then predict BP from the reconstructed ECG using the same BiLSTM model. Using an identical BP prediction model ensures a controlled, apples-to-apples comparison between pipelines.

\subsection{Generators (PPG$\rightarrow$ECG)}

We evaluate four generator architectures for reconstructing ECG from PPG:

\textbf{U-Net} \cite{ronneberger2015unet}: Convolutional encoder-decoder with skip connections and MaxPooling downsampling (1.23M parameters).

\textbf{Wave-U-Net} \cite{stoller2018waveunet}: Uses learnable strided convolutions and large kernels to preserve temporal precision critical for ECG morphology (1.83M parameters).

\textbf{Transformer} \cite{vaswani2017attention}: Uses self-attention with positional encoding to capture long-range temporal dependencies (0.17M parameters).

\textbf{CycleGAN} \cite{zhu2017cyclegan}: Uses adversarial and cycle-consistency losses to enforce bidirectional signal consistency (9.45M parameters).

Model and pipeline complexity are summarized in Table~\ref{tab:model_comparison}.

\begin{table}[htbp]
\caption{Model and Pipeline Complexity}
\begin{center}
\begin{tabular}{|l|c|c|c|}
\hline
\textbf{Model} & \textbf{Params} & \textbf{Train} & \textbf{Infer} \\
\hline
\multicolumn{4}{|c|}{\textit{Generators (PPG$\rightarrow$ECG)}} \\
\hline
U-Net & 1.23M & 8.2m & 12ms \\
Wave-U-Net & 1.83M & 11.5m & 15ms \\
Transformer & 0.17M & 14.3m & 18ms \\
CycleGAN & 9.45M & 45.7m & 25ms \\
\hline
\multicolumn{4}{|c|}{\textit{Direct Pipeline (PPG$\rightarrow$BP)}} \\
\hline
\textbf{BiLSTM} & \textbf{0.15M} & \textbf{3.1m} & \textbf{5ms} \\
\hline
\multicolumn{4}{|c|}{\textit{ECG-Mediated Pipelines (PPG$\rightarrow$ECG$\rightarrow$BP)}} \\
\hline
Wave-U-Net + BiLSTM & 1.98M & 14.6m & 20ms \\
CycleGAN + BiLSTM & 9.60M & 48.8m & 30ms \\
\hline
\end{tabular}
\label{tab:model_comparison}
\end{center}
\end{table}

\subsection{BP Prediction Model (PPG$\rightarrow$BP and ECG$\rightarrow$BP)}

Both the Direct Pipeline and ECG-Mediated Pipelines use the same BiLSTM architecture to predict SBP and DBP, ensuring a fair comparison. The model consists of a BiLSTM layer (128 units, dropout 0.3), followed by global average pooling and two fully connected layers (64 units with ReLU and dropout 0.2, and final linear output). The model has 149,698 parameters.

\subsection{Training Configuration}

Generators (PPG$\rightarrow$ECG) are trained using Adam optimization (learning rate $10^{-3}$; $2 \times 10^{-4}$ for CycleGAN), mean squared error loss, and early stopping (patience = 3). BP prediction models in both Direct and ECG-Mediated Pipelines are trained using identical settings with batch size 64. All experiments are conducted on a Tesla T4 GPU.

\section{Evaluation Metrics}

\textbf{Generators (PPG$\rightarrow$ECG)} are evaluated using mean absolute error (MAE), mean squared error (MSE), root mean squared error (RMSE), Pearson correlation, and R-peak detection accuracy using the Pan–Tompkins algorithm, assessing reconstruction fidelity, morphology preservation, and temporal precision.

\textbf{Direct Pipeline (PPG$\rightarrow$BP)} and \textbf{ECG-Mediated Pipelines (PPG$\rightarrow$ECG$\rightarrow$BP)} are evaluated using mean absolute error (MAE) and standard deviation (SD) between predicted and ground-truth ABP-derived SBP and DBP. Performance is assessed according to British Hypertension Society (BHS) standards: Grade A (MAE $\leq$ 5 mmHg, SD $\leq$ 8 mmHg), Grade B (MAE $\leq$ 10 mmHg, SD $\leq$ 15 mmHg), and Grade C (MAE $\leq$ 15 mmHg, SD $\leq$ 20 mmHg).

\section{Results}

\subsection{Baseline Performance}

Table~\ref{tab:baseline} shows baseline performance using traditional and early deep learning approaches. Linear regression based on hand-crafted features performs poorly (Grade D), confirming the limitations of feature-engineered pipelines. An LSTM baseline improves performance (Grade C), but remains below clinical standards.

\begin{table}[htbp]
\caption{Baseline BP Estimation Performance (MAE in mmHg)}
\begin{center}
\begin{tabular}{|l|c|c|c|c|c|}
\hline
\textbf{Model} & \textbf{SBP} & \textbf{SBP} & \textbf{DBP} & \textbf{DBP} & \textbf{BHS} \\
 & \textbf{MAE} & \textbf{SD} & \textbf{MAE} & \textbf{SD} & \textbf{Grade} \\
\hline
Linear Regression & 20.86 & 23.93 & 12.66 & 13.76 & D \\
LSTM \cite{kachuee2018} & 9.52 & 13.12 & 6.17 & 10.40 & C \\
\hline
\end{tabular}
\label{tab:baseline}
\end{center}
\end{table}

\subsection{Generator Performance (PPG$\rightarrow$ECG)}

Table~\ref{tab:ecg_quality} compares generators that reconstruct ECG from PPG. Wave-U-Net achieves the best performance across all reconstruction and physiological metrics, including lowest MAE (0.142), highest R-peak detection accuracy (94.7\%), and strongest morphology correlation (0.891). Compared to the baseline U-Net, this corresponds to a 24\% reduction in reconstruction error, indicating substantially improved temporal and morphological fidelity.

This improvement is primarily attributable to architectural design. Wave-U-Net replaces fixed MaxPooling with learnable strided convolutions, enabling adaptive downsampling that preserves physiologically relevant timing information. Its larger receptive fields (kernel size 15) capture longer cardiac dynamics, while skip connections maintain fine-grained temporal detail across scales. 

Transformer-based generators did not outperform convolutional models, likely because the input sequences are short (256 samples) and ECG morphology is governed primarily by local temporal structure. In this regime, convolutional architectures efficiently capture the full physiological context through hierarchical receptive fields, reducing the benefit of global self-attention.

CycleGAN achieved moderate performance but required significantly higher complexity and training time (Table~\ref{tab:model_comparison}), reflecting the overhead of adversarial training without proportional gains in reconstruction accuracy.

\begin{table}[htbp]
\caption{PPG-to-ECG Generation Quality}
\begin{center}
\begin{tabular}{|c|c|c|c|c|c|}
\hline
\textbf{} & \textbf{} & \textbf{} & \textbf{} & \textbf{R-Peak} & \textbf{Morph.} \\
\textbf{Model} & \textbf{MAE} & \textbf{MSE} & \textbf{RMSE} & \textbf{(\%)} & \textbf{Corr.} \\
\hline
U-Net & 0.187 & 0.0512 & 0.226 & 91.3 & 0.847 \\
\hline
\textbf{Wave-U-Net} & \textbf{0.142} & \textbf{0.0298} & \textbf{0.173} & \textbf{94.7} & \textbf{0.891} \\
\hline
Transformer & 0.165 & 0.0387 & 0.197 & 92.8 & 0.863 \\
\hline
CycleGAN & 0.158 & 0.0351 & 0.187 & 93.2 & 0.874 \\
\hline
\end{tabular}
\label{tab:ecg_quality}
\end{center}
\end{table}

\begin{figure}[h!]
    \centering
    \includegraphics[width=0.90\linewidth]{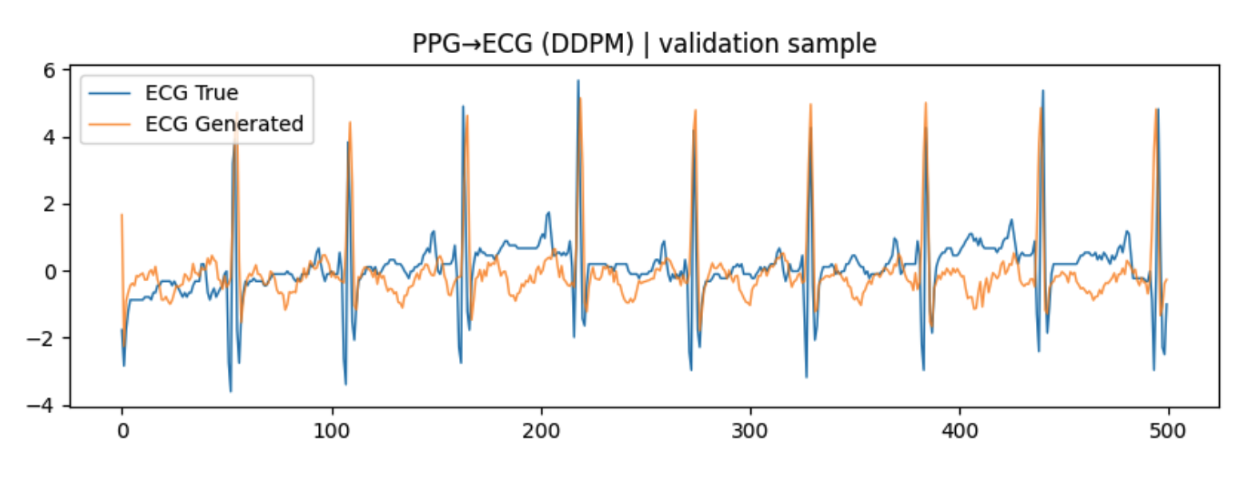}
    \caption{Comparison between ground-truth ECG and the ECG generated by the diffusion-based PPG$\rightarrow$ECG model on a validation sample.}
    \label{fig:ppg2ecg_visual}
\end{figure}

Figure~\ref{fig:ppg2ecg_visual} illustrates representative generated ECG signals. The generators successfully reconstruct overall waveform morphology, including QRS complexes and cardiac rhythm. However, small temporal misalignments in R-peak timing remain, which introduce downstream prediction error. These results demonstrate that while generators can reconstruct physiologically plausible ECG, reconstruction imperfections may affect subsequent BP estimation.

\subsection{Blood Pressure Estimation}

\renewcommand{\arraystretch}{1.25}
\begin{table*}[t]
\caption{Blood Pressure Estimation Performance Comparison  (MAE in mmHg)}
\centering
\begin{tabular}{|p{5.2cm}|p{1.8cm}|p{1.6cm}|p{1.6cm}|p{1.6cm}|p{1.6cm}|p{1.2cm}|}
\hline
\textbf{Model / Pipeline} & \textbf{Input} & \textbf{SBP MAE} & \textbf{SBP SD} & \textbf{DBP MAE} & \textbf{DBP SD} & \textbf{Grade} \\
\hline

\multicolumn{7}{|l|}{\textit{Upper Bound Baseline}} \\
\hline
Real ECG $\rightarrow$ BiLSTM & temporal & 4.23 & 6.89 & 3.87 & 6.54 & A \\
\hline

\multicolumn{7}{|l|}{\textit{Direct Pipeline (PPG$\rightarrow$BP)}} \\
\hline
\textbf{PPG $\rightarrow$ BiLSTM} & temporal & \textbf{4.82} & \textbf{7.65} & \textbf{4.31} & \textbf{7.12} & \textbf{A} \\
\hline

\multicolumn{7}{|l|}{\textit{ECG-Mediated Pipelines (PPG$\rightarrow$ECG$\rightarrow$BP)}} \\
\hline
PPG $\rightarrow$ U-Net $\rightarrow$ BiLSTM & temporal & 6.94 & 9.87 & 5.73 & 8.91 & B \\
\hline
PPG $\rightarrow$ Wave-U-Net $\rightarrow$ BiLSTM & temporal & 5.38 & 8.24 & 4.89 & 7.68 & B \\
\hline
PPG $\rightarrow$ Transformer $\rightarrow$ BiLSTM & temporal & 6.12 & 9.13 & 5.21 & 8.35 & B \\
\hline
PPG $\rightarrow$ CycleGAN $\rightarrow$ BiLSTM & temporal & 5.76 & 8.67 & 5.04 & 8.02 & B \\
\hline
PPG $\rightarrow$ Diffusion $\rightarrow$ CNN (v1) & spectral & 19.0 & -- & 19.0 & -- & -- \\
\hline
PPG $\rightarrow$ Diffusion $\rightarrow$ CNN (v2) & spectral & 10.70 & -- & 12.85 & -- & -- \\
\hline

\multicolumn{7}{|l|}{\textit{Mixed-Input Pipelines}} \\
\hline
PPG + ECG $\rightarrow$ CNN & spectral & 10.7 & -- & 13.0 & -- & -- \\
\hline
PPG + ECG $\rightarrow$ BiLSTM & spectral & 12--13 & -- & 14--15 & -- & -- \\
\hline

\end{tabular}
\label{tab:main_results}
\end{table*}

Table~\ref{tab:main_results} presents the main comparison between the Direct Pipeline (PPG$\rightarrow$BP) and ECG-Mediated Pipelines (PPG$\rightarrow$ECG$\rightarrow$BP).

The Direct Pipeline achieves British Hypertension Society Grade A performance, with MAE: 4.82/4.31 mmHg and SD: 7.65/7.12 mmHg for SBP/DBP. In contrast, all ECG-Mediated Pipelines achieve only Grade B performance, despite using state-of-the-art generators.

Among ECG-Mediated Pipelines, Wave-U-Net + BiLSTM performs best (MAE: 5.38/4.89 mmHg), but remains inferior to the Direct Pipeline. This corresponds to an 11.6\% higher SBP error and 13.5\% higher DBP error. This difference is statistically significant ($p < 0.001$, paired t-test). 

Using real ECG as input achieves Grade A performance, confirming that the performance degradation arises from the ECG generation step rather than the BP predictor itself.

\subsection{Spectral Domain Experiments}

We further evaluated Direct and ECG-Mediated Pipelines using spectral features derived via FFT. Direct spectral prediction achieves MAE: 10.7/13.0 mmHg for SBP/DBP. In contrast, ECG-Mediated Pipelines using diffusion-based generators perform substantially worse, with errors up to MAE: 19.0 mmHg.

We implement a conditional 1D diffusion model (DDPM) to generate ECG spectra from PPG spectra. The model is trained using a time-conditioned noise prediction objective, with both end-to-end and two-stage training strategies. However, both approaches underperform compared to direct prediction.

\subsection{Statistical Validation}

We performed paired t-tests on 217,456 test samples to compare the Direct Pipeline and ECG-Mediated Pipelines. The Direct Pipeline shows significantly lower error for both SBP ($t(217455) = 24.7$, $p < 0.001$, Cohen's $d = 0.168$) and DBP ($t(217455) = 22.3$, $p < 0.001$, Cohen's $d = 0.151$).

Bland–Altman analysis further confirms improved agreement for the Direct Pipeline, with narrower limits of agreement compared to the best ECG-Mediated Pipeline (Wave-U-Net). For SBP, the limits are $\pm 15.2$ mmHg versus $\pm 19.8$ mmHg, and for DBP, $\pm 14.1$ mmHg versus $\pm 17.6$ mmHg. Both pipelines show negligible systematic bias.

\subsection{Demographic Generalization}

Table~\ref{tab:demographics} shows performance across patient subgroups. The Direct Pipeline (PPG$\rightarrow$BP) consistently outperforms the best ECG-Mediated Pipeline (Wave-U-Net) across all demographic groups.

Performance slightly degrades in older and hypertensive patients, likely due to increased vascular variability. Nevertheless, the Direct Pipeline maintains Grade A performance across all subgroups, demonstrating robust generalization.

\begin{table}[htbp]
\caption{Performance Across Patient Subgroups (MAE in mmHg)}
\begin{center}
\begin{tabular}{|c|c|c|c|c|}
\hline
\textbf{Subgroup} & \multicolumn{2}{c|}{\textbf{Direct}} & \multicolumn{2}{c|}{\textbf{ECG-Mediated Pipeline}} \\
\cline{2-5}
 & \textbf{SBP} & \textbf{DBP} & \textbf{SBP} & \textbf{DBP} \\
\hline
Age $<$ 50 & 4.67 & 4.18 & 5.21 & 4.76 \\
\hline
Age $\geq$ 50 & 4.91 & 4.39 & 5.48 & 4.97 \\
\hline
Male & 4.79 & 4.28 & 5.34 & 4.85 \\
\hline
Female & 4.86 & 4.35 & 5.43 & 4.93 \\
\hline
Normal BP & 3.98 & 3.76 & 4.67 & 4.21 \\
\hline
Hypertensive & 5.34 & 4.67 & 6.02 & 5.38 \\
\hline
\end{tabular}
\label{tab:demographics}
\end{center}
\end{table}

\section{Discussion}

\subsection{Why the Direct Pipeline Outperforms ECG-Mediated Pipelines}

\textbf{Information bottleneck}: ECG generation introduces an intermediate representation that does not preserve all BP-relevant information present in PPG. PPG directly reflects peripheral hemodynamics, including pulse amplitude, waveform morphology, and vascular compliance, which are physiologically linked to blood pressure. In contrast, ECG reflects electrical cardiac activity and shows very weak correlation with ABP ($r = 0.018$, Table~\ref{tab:correlation}). As a result, translating PPG to ECG discards information that is useful for BP estimation but not encoded in ECG morphology.

\textbf{Error propagation}: The ECG-Mediated Pipeline introduces two sequential sources of error: ECG generation and BP estimation. These errors accumulate, such that
\[
\text{Error}_{\text{total}} = \text{Error}_{\text{generation}} + \text{Error}_{\text{estimation}}.
\]
Even the best generator (Wave-U-Net, MAE = 0.142, Table~\ref{tab:ecg_quality}) introduces reconstruction error, which degrades downstream BP prediction. In contrast, the Direct Pipeline optimizes BP estimation end-to-end, avoiding intermediate reconstruction errors.

\textbf{Objective misalignment}: ECG generation models are trained to minimize signal reconstruction loss (MSE), which does not directly optimize BP estimation. Small timing shifts in generated ECG (e.g., R-peak misalignment, Fig.~\ref{fig:ppg2ecg_visual}) have minimal impact on reconstruction metrics but can degrade BP-relevant features. In contrast, the Direct Pipeline optimizes the BP prediction objective directly, resulting in better alignment between training loss and clinical outcome.

\subsection{Clinical Implications}

These results support direct PPG-to-BP estimation as the preferred approach for wearable systems.

The Direct Pipeline achieves Grade A performance, while all ECG-Mediated Pipelines achieve only Grade B (Table~\ref{tab:main_results}). In addition to higher accuracy, the Direct Pipeline is computationally simpler, requiring only a single model and fewer parameters (0.15M vs up to 9.6M, Table~\ref{tab:model_comparison}). This reduces inference time, memory footprint, and power consumption, which are critical constraints for wearable devices.

Grade A performance meets British Hypertension Society standards and supports the feasibility of continuous cuffless BP monitoring using wearable PPG sensors.

\subsection{Limitations}

This study uses the MIMIC-III dataset, which contains ICU patients. Validation in ambulatory and healthy populations is required to confirm generalization to wearable settings. The signals are high-quality clinical recordings, whereas wearable devices introduce motion artifacts and sensor variability.

BP ground truth was obtained from arterial lines, which provide higher precision than cuff measurements used in consumer devices. The models operate on short signal windows and do not evaluate long-term continuous monitoring. In addition, the models do not include subject-specific calibration, which may further improve accuracy.

\subsection{Future Directions}

Future work should evaluate direct PPG-based BP estimation in real-world wearable settings. Multi-task learning approaches that jointly predict BP and other physiological signals may improve efficiency. Personalization using small calibration datasets may further improve accuracy. Prospective validation on wearable devices during daily activities is necessary to confirm clinical applicability.

\section{Conclusions}

Many prior cuffless blood pressure estimation approaches assume that reconstructing ECG from wearable PPG provides a stronger physiological representation for downstream prediction. In this work, we directly tested this assumption through both physiological analysis and controlled modeling experiments. Our correlation analysis on 1.74M segments from 3,127 patients shows that PPG is substantially more strongly coupled to arterial blood pressure than ECG ($|r|=0.247$, $p<0.001$ vs. $r=0.018$, $p=0.187$), questioning the necessity of ECG as an intermediate representation.

Motivated by this finding, we conducted a systematic comparison between Direct Pipeline (PPG$\rightarrow$BP) and ECG-Mediated Pipelines (PPG$\rightarrow$ECG$\rightarrow$BP) using multiple state-of-the-art generative models. The Direct Pipeline achieves British Hypertension Society Grade A performance (MAE: 4.82/4.31 mmHg for SBP/DBP), while all ECG-Mediated Pipelines achieve only Grade B accuracy, despite high-quality ECG reconstruction. Statistical analysis confirms this improvement is significant ($p < 0.001$). These results demonstrate that intermediate ECG generation may introduce an information bottleneck and error propagation that degrade downstream prediction, whereas direct end-to-end learning preserves BP-relevant hemodynamic information present in PPG.

Among the evaluated generators, Wave-U-Net achieves the best ECG reconstruction quality, confirming the effectiveness of learnable multi-scale architectures for physiological signal synthesis when ECG reconstruction itself is required. However, ECG reconstruction does not improve BP estimation accuracy.

Overall, this study provides physiological and empirical evidence that accurate cuffless blood pressure estimation can be achieved directly from wearable PPG signals without intermediate ECG reconstruction. 

\section*{Acknowledgment}

This work was conducted as part of the Carnegie Mellon University 11-785 Deep Learning course (Fall 2025): \url{https://deeplearning.cs.cmu.edu/S25/index.html}.

% \vspace{12pt}
% \color{red}
% IEEE conference templates contain guidance text for composing and formatting conference papers. Please ensure that all template text is removed from your conference paper prior to submission to the conference. Failure to remove the template text from your paper may result in your paper not being published.

\end{document}